\documentclass[a4paper, 10pt, conference]{ieeeconf}      % Use this line for a4 paper

\IEEEoverridecommandlockouts                              % This command is only needed if 
\usepackage{cite}
\usepackage{graphicx}
\title{\LARGE \bf
Autonomous Rollator: A Case Study in the Agebots Project
}

\author{Jonas Frei$^{1}$, Anina Havelka$^{2}$, Markus Wüst$^{1}$, Einar Nielsen$^{1}$, Andreas Ziltener$^{2}$ and Katrin S. Lohan$^{1,3}$% <-this % stops a space
\thanks{$^{1}$
        Eastern Switzerland University of Applied Sciences, Buchs, Switzerland
        {\tt\small jonas.frei@ost.ch}}%
\thanks{$^{2}$
University of Applied Sciences of the Grisons, Chur, Switzerland 
        }%
\thanks{$^{3}$
Heriot-Watt University, Edinburgh, United Kingdom  
        }%        
}

\begin{document}

\maketitle
\thispagestyle{empty}
\pagestyle{empty}

%%%%%%%%%%%%%%%%%%%%%%%%%%%%%%%%%%%%%%%%%%%%%%%%%%%%%%%%%%%%%%%%%%%%%%%%%%%%%%%%
\begin{abstract}

In this paper, we present an iterative development process for a functional model of an autonomous, location-orienting rollator. An interdisciplinary team was involved in the development, working closely with the end-users. This example shows that the design thinking method is suitable for the development of frontier technology devices in the care sector.   

\end{abstract}
\hspace{10pt}


%%%%%%%%%%%%%%%%%%%%%%%%%%%%%%%%%%%%%%%%%%%%%%%%%%%%%%%%%%%%%%%%%%%%%%%%%%%%%%%%
\section{Introduction}
 The Agebots project "Implementation of service robotics in the field of elderly care", is an interdisciplinary, federally funded project following a human-centered and iterative approach to support the elderly care with service robotic solutions. It focuses on the needs and challenges of the caregivers and the elderly in these care facilities. Different design thinking methods are used to elicit those needs and challenges, including multiple interviews with potential end-users, observations, shadowings, personae, rapid prototyping, laboratory, and field testing. Considering the different dimensions, i.e., the human, technology, business, and law aspects, prevents a one-sided, technical driven approach and promotes the implementation of requirement-based solutions.
The increase in the elderly population in Switzerland will reach an expected share of over 27\% of the 65+ population group (2.8 million) by 2050. This will cause major challenges, especially for retirement and nursing homes \cite{fur2017bevolkerung, suisse2015zukunft}. This ageing of society, in combination with work overload, changing care needs, lack of time, human errors, and economic pressure lead to a decline in the quality of care. To counteract this decline and ensure the satisfaction of the elderly and caregivers, the care facilities are dependent on innovative and financially feasible solutions \cite{klein2018robotik}. These innovative solutions should ensure a high-quality, and efficient care for the elderly in the future. Furthermore, the solutions should relieve the caregivers to give them more time for interpersonal activities. Considering these developments, service robotics might be the solution.

\section{The Agebots Project}
The Agebots project is structured in such a way that the robotic solutions are tailored to the challenges and needs of the end-users and are developed and tested in two iterations (cf. Table \ref{table_1}). The focus of the first testing was the technical feasibility and implementation. Here the robotic solutions were tested first with non-users. In the second Lab Test the robot systems were tested with end-users, i.e., elderly and caregivers of the retirement and nursing home. The focus of the second testing was on acceptance in terms of user-friendliness, readiness to use and personal impression. For each of the Lab Tests, a concept was created with specific observation criteria. The application fields could be tested several times by different users resulting in a variety of feedbacks and inputs. In addition, observations were carried out which allowed for identification of interaction difficulties with the robots. The feedback was implemented onsite (laboratory), allowing to continue the testing with the improved functionalities. Specifically, this iterative approach helped to implement and test the end-users' requirements directly, so that their wishes could be considered in the best possible way. The range of applications, which were tested in the two Lab Tests, were reduced to the most feasible or most promising ones throughout the testing phases.
\begin{table}[h]
\caption{Project Phases and general Results}
\label{table_1}
\begin{center}
\begin{tabular}{|p{2cm}|p{6cm}|}
\hline
Project phases & Results\\
\hline
\hline
Understanding & One hundred forty-four challenges for the caregivers and the elderly were identified.\\
\hline
Ideation & Fifteen application fields with 170 ideas for caregivers and ten application fields with 88 ideas for elderly were generated.\\
\hline
Technology & Seven robotic systems were procured, and functional tests were conducted.\\
\hline
Lab test 1 & Feasibility tests carried out with non-users (students, employees from the retirement and nursing home, and the universities), and first legal analysis was conducted. 25 applications were reduced to 12.\\
\hline
Iteration 1 & Application development of the robot systems based on findings of lab test 1. Research framework for the evaluation of the qualitative and quantitative data using a relational database was created and filled with insights.\\
\hline
Lab test 2 & Usability and acceptance tests with the caregivers and elderly of the specific retirement and nursing home in a laboratory setting. A business plan for the development of new markets for the implementation partner (here robot manufacturer) was elaborated.\\
\hline
Field testing (planned) & Testing of legally compliant, desirable, feasible, and economical application in real-world environments in the retirement and nursing home.\\
\hline
Marketing (planned) & "Sale" and distribution of the service robot systems in German-speaking countries by the main implementation partner (robot manufacturer).\\
\hline
\end{tabular}
\end{center}
\end{table}

\section{Companion Challenge}
One of the 144 identified challenges was the Companion Challenge which will now be discussed. Caregivers need to escort the elderlies several times a day to different locations within the retirement and nursing homes. A necessary but very time consuming activity for the caregivers, identified in the literature as the most requested feature to maintain the independence of the elderly \cite{hebesberger2015staff, pineau2003towards}. Therefore, the goal of this project is to find innovative assistance systems and robotic solutions, which can relieve or support the caregivers in this task, and preserve the elderlies independence.

\section{The Autonomous Rollator Design Process}
In this section we will discuss three iterations of the development of our autonomous rollator as an escorter robot.
\subsection{First iteration}
\begin{figure}
    \centering
    \includegraphics[width=\linewidth]{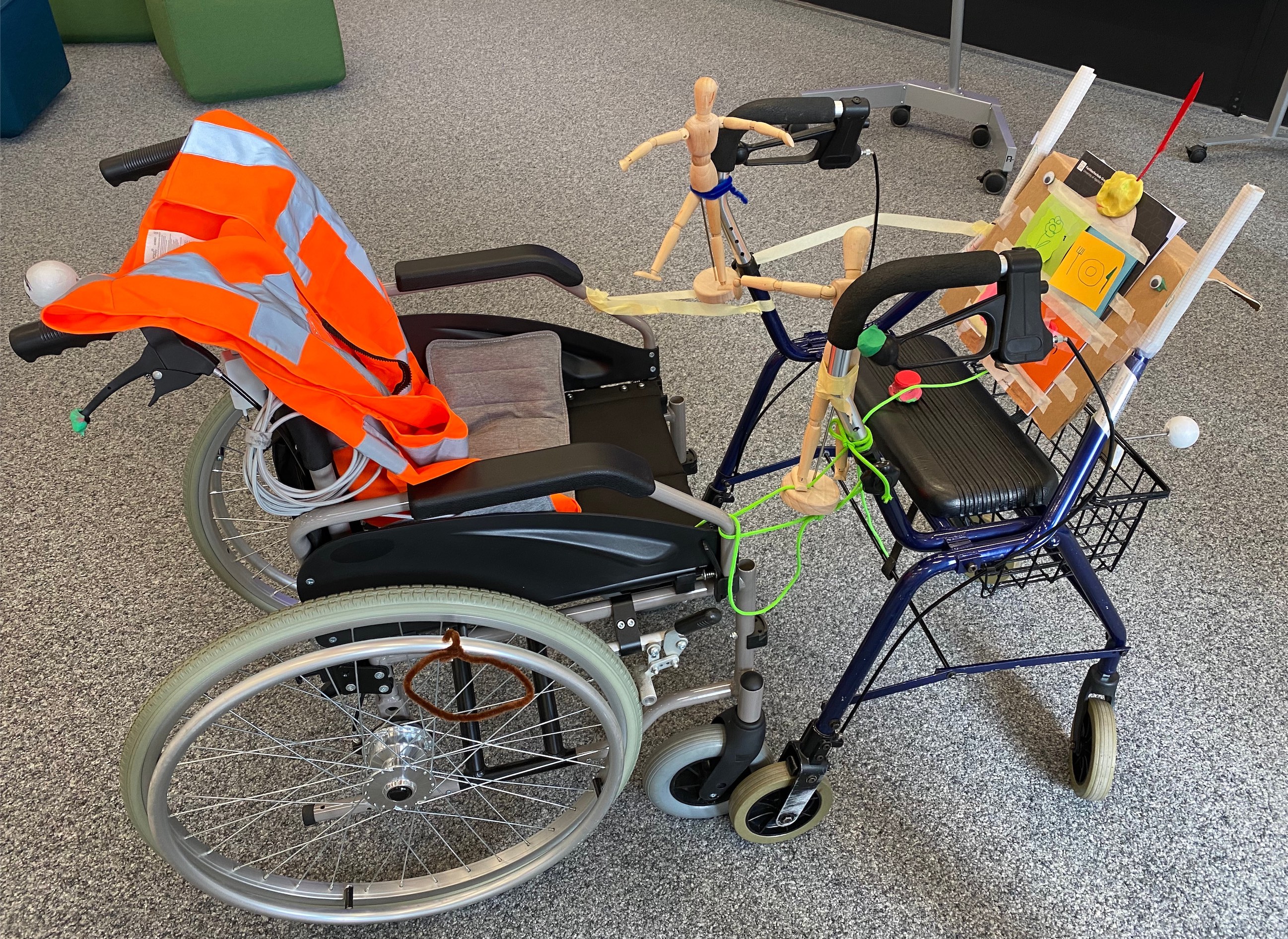}
    
  \caption{Low-fidelity prototype of the rollator}
 \label{fig:experiment_1}
\end{figure}

In the case of the Companion Challenge, one result of a creativity workshop was a rollator combined with a wheelchair. This first low-fidelity prototype was made of cardboard, cord, post-its and drawings (cf. Figure \ref{fig:experiment_1}).
At this workshop, the technical requirements for a possible solution were discussed and defined with the end-users, which were then procured, provided, and put into operation.

\subsection{Second iteration}
In the first Lab Test (see Table \ref{table_1}), the robot systems and the evaluated application were tested with different non-users such as students and employees of universities as well as employees of the retirement and nursing home.

During the technical development, the following requirements for the Companion System were identified in accordance with the findings of \cite{hawes2017strands,pineau2003towards}:
\begin{itemize}
\item[a)]	Determining the position and orientation in space
\item[b)]	Generating a map, annotating important locations, and restricting areas
\item[c)]	Path planning considering temporary obstacles and physical variability
\item[d)]	Ensuring long-term autonomy of a system 
\item[e)]	Determining the current floor
\item[f)]	Communicating with, monitoring of, and adapting to the user
\item[g)]	Ease of operation
\end{itemize}
Based on these technical requirements we adjusted the rollator from ello to escort the elderly autonomously. 
In the following, we will discuss two iterations of modification of this rollator to become an autonomous robot.

\begin{figure}
    \centering
    \includegraphics[width=\linewidth]{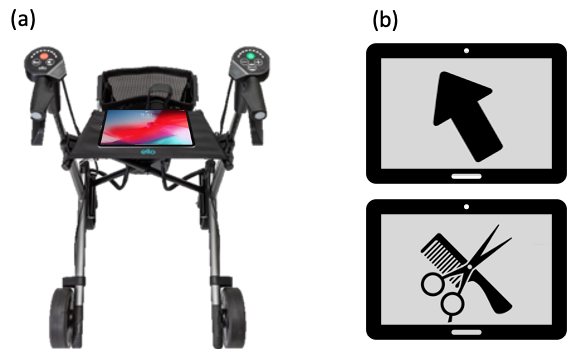}
    
  \caption{(a) Picture of the prototype in the first iteration. (b) Graphical user interface example used in the first iteration.}
 \label{fig:experiment_2}
\end{figure}
As a second iteration, we used the Wizard of Oz (WOZ) approach \cite{dahlback1993wizard}. The aim was to get an initial feeling for the requirement f) and the associated needs of the system. For this purpose, a tablet was mounted on a rollator. By means of the tablet, the wizard interacted visually and acoustically with the elderly (cf. Figure \ref{fig:experiment_2}).

The results achieved with the second prototype proved that communication via tablet works in principle. The WOZ approach allowed the wizard to interact with and actively influence the behavior of the elderly. Though, implementing this kind of interaction and influence on a computer and realizing an autonomous accompanying system with it is challenging. Additionally, it is difficult for the elderly to concentrate on the path and read the information on the tablet at the same time. This is especially problematic for people who are no longer sufficiently mobile. In summary, a communication tool other than the tablet still needs to be found.

\subsection{Third iteration}
\begin{figure}
    \centering
    \includegraphics[width=\linewidth]{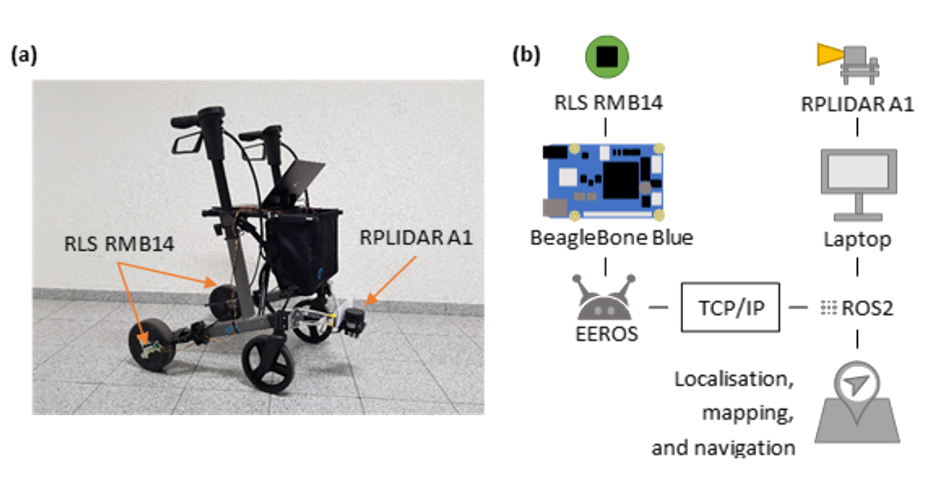}
    
  \caption{(a) Picture of the prototype of the third experiment. (b) System overview of the third experiment.}
 \label{fig:experiment_3}
\end{figure}

For the second Lab Test (cf. Table \ref{table_1}), the rollator was rebuilt, as can be seen in Figure \ref{fig:experiment_3}, to tackle the previously mentioned requirements a) to c). The aim was to shift from the WOZ approach to an autonomous system and to become better acquainted with the possible applications. The rollator was equipped with two RLS RMB14 encoders connected to a BeagleBone blue single-board computer and a RPLIDAR A1 light detection and ranging sensor connected to a laptop. The BeagleBone blue and the laptop communicate over TCP/IP. Using odometry the position and orientation of the rollator is estimated. These calculations are done in the EEROS framework, and the results are sent to the ROS2 software stack \cite{thomas2014next}. To perform simultaneous localization and mapping (SLAM) the slam toolbox \cite{macenski2019use,macenski2021slam}, and for the navigation the Nav2 project \cite{macenski2020marathon} is used.
Caregivers were able to test the third prototype in a replicated elderly residence. Their feedback was positive, and they see potential in the application. From a technical point of view, the increasing computational effort required by SLAM with the size of the mapped area is a limiting factor. Another problem was that the rollator localization sometimes suddenly switched its location to the other side of a wall when it came too close to the wall. However, this was only observed in the replicated retirement and nursing home and may be understood as the walls are made of thin wooden panels. Nevertheless, this caused the map to become unusable and therefore the application had to be restarted. A possible solution would be to switch to a static map and only use SLAM for the initial recording of the map. Overall, further research is required to improve the prototype and tackle the remaining requirements.
\section{Discussion and Conclusion}
The human-centered approach in the analysis and implementation phase of technical assistance systems offers the possibility to align solutions directly with the needs of end-users. This allows corrections in the design process before the robotic solutions are implemented in the real environment. Through repeated iterative testing, prototypes are continuously adapted and improved.

Well-defined requirements for a companion system were identified and three prototypes were successfully developed to gain experience in dealing with these challenges. Problems in communicating with the elderly via a display were identified using the WOZ approach. The technical feasibility of localisation and navigation with a walker was demonstrated and the solution was evaluated by caregivers.

The results obtained in the second iteration showed that further research is needed in the identified challenges f) communicating with, monitoring of, and adapting to the user, as well as g) ease of operation. One possible solution, which will be investigated in a follow up study, is the implementation of force guided navigation. Like a lane departure warning system, the aim is to subtly point the user in the right direction. Ultimately leading to an intuitive communication that leaves the user's visual senses free for other tasks.

Although some software improvements are still needed to ensure the long-term autonomy, the prototype built in the third iteration demonstrated that escorting is possible with simple and inexpensive sensors. Furthermore, the positive oral feedback from caregivers about the usefulness of such a system is an important factor in deciding that the system should be further developed in a subsequent study.

Two steps are planned for the further investigation. In a first step, it is planned to fix the still open communication issues and the issue with the generated map, and to integrate the requirements -- d), e) and g) -- that have been neglected so far. In a second step, a manufacturer of rollators is to be sought so that a prototype ready for series production can be developed in a separate application-oriented research project. In both steps, tests in the laboratory and in real environments are to be carried out.

\bibliographystyle{IEEEtran}
\bibliography{example}

\end{document}